\documentclass[letterpaper]{article}
\usepackage[preprint]{aaai2027}

\usepackage[utf8]{inputenc}
\usepackage[T1]{fontenc}
\usepackage[hyphens]{url}
\usepackage{graphicx}
\usepackage{natbib}
\usepackage{caption}
\usepackage{booktabs}
\usepackage{amsfonts}
\usepackage{amsmath}
\usepackage{amssymb}
\usepackage{microtype}
\usepackage{xcolor}
\usepackage{subcaption}
\usepackage{algorithm}
\usepackage{algorithmic}
\usepackage{multirow}

\newcommand{\ours}{\textsc{$\varepsilon$-MemEvo}}
\newcommand{\ada}{\textsc{AdaEvolve}}
\newcommand{\topk}{\textsc{TopK}}

\title{$\varepsilon$-MemEvo: Adaptive Cross-Task Memory Transfer\\for LLM Program Evolution}
\author{Aofan Liu, Song Shiyuan, Qi Yiyan\corresponding}
\affiliations{}

\begin{document}

\maketitle
\enlargethispage{-18pt}

\begin{abstract}
LLM-based program evolution systems such as FunSearch and AlphaEvolve have shown strong ability to discover novel algorithms, but typically optimize each task in isolation, discarding search experience after completion. We introduce \ours{}, a framework for \emph{cross-task knowledge transfer} in LLM program evolution. \ours{} stores prior experience as \emph{task-agnostic tactic memories}: compact natural-language summaries of successful algorithmic strategies rather than raw code, enabling transfer across tasks with different APIs and evaluators. To avoid negative transfer from semantically mismatched memories, \ours{} uses an adaptive injection gate that decides whether retrieved memories should be injected, and at what intensity.
We evaluate \ours{} on 8 diverse optimization benchmarks spanning mathematical optimization and systems engineering, using a content-level Leave-One-Out protocol that excludes target-task memory entries. On the primary GPT-5 backbone, \ours{} improves AUCC over AdaEvolve on all 8 tasks, with a mean relative gain of $+8.7\%$, and improves early-stage convergence by $+9.4\%$ on average. Ablations show that naive memory injection can fail catastrophically, while adaptive gating remains safe across all five ablation tasks. The data-updated posterior is interpretable in observed states: it favors \emph{skip} during improving search and shifts from \emph{skip} to \emph{hint} across early and late plateaus. These gains incur $<\!1\%$ computational overhead.
\end{abstract}

\section{Introduction}

Large language models (LLMs) have emerged as powerful engines for automated algorithm discovery. Systems like FunSearch~\citep{romera2024funsearch} and AlphaEvolve~\citep{novikov2025alphaevolve} use LLMs to iteratively generate, evaluate, and refine programs, achieving breakthroughs in combinatorial optimization, matrix multiplication, and scientific computing. More recent work such as AdaEvolve~\citep{cemri2026adaevolve} introduces adaptive island models and paradigm breakthroughs to further improve search diversity.

Despite these advances, a fundamental limitation persists: \textbf{every task starts from scratch}. When an LLM evolution system finishes one optimization task and moves to a related task, it retains nothing---no memory of which strategies worked, no intuition about search structure, no reusable heuristics. This ``amnesic'' behavior leads to redundant exploration across related tasks, slow cold-start convergence, and wasted API budget.

\begin{figure}[t]
    \centering
    \includegraphics[width=0.95\columnwidth]{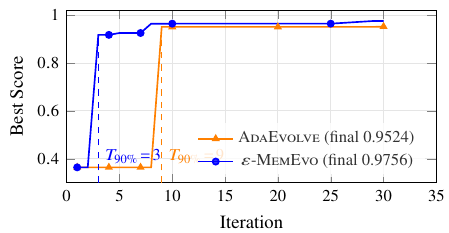}
    \caption{\textbf{Motivating example.} On circle packing after solving a related geometric task, \ours{} retrieves a geometric arrangement heuristic from memory and reaches 90\% of its final score in 3 iterations (vs.\ 9 for AdaEvolve). The Tactic Memory Bank stores strategy summaries (not code), enabling transfer across tasks with different APIs and metrics.}
    \label{fig:motivation}
\end{figure}

Consider a concrete scenario: after optimizing a geometric packing problem, the system has learned that ``structured grid layouts with symmetric offsets minimize overlap.'' This geometric intuition can also benefit a related packing task, yet existing systems cannot leverage it. Human researchers naturally build on prior experience---why shouldn't LLM evolution systems?
We propose \ours{} (\textbf{$\varepsilon$-Mem}ory-augmented \textbf{Evo}lution), a framework for cross-task knowledge transfer in LLM-based program evolution. \ours{} makes two methodological moves tailored to evolutionary coding:

\begin{itemize}
    \item \textbf{Content-level transfer} via a \emph{Tactic Memory Bank} that stores LLM-extracted natural-language strategy summaries (not raw code) from completed tasks, enabling transfer across heterogeneous APIs and evaluators.
    \item \textbf{Strategy-level gating} via an \emph{Adaptive Injection Gate} that treats memory usage as an adaptive gating problem whose primary role is to avoid negative transfer: it learns when retrieved memories should be withheld, softly suggested, or strongly imposed based on search state.
\end{itemize}

The key insight behind \ours{} is that \emph{cross-task transfer in evolutionary coding is a safety problem, not a retrieval problem alone}. Na\"ively injecting cross-task strategies every iteration can cause catastrophic negative transfer---in our ablation, two of five tasks see the search fail to produce \emph{any} score-improving program under always-inject, and the same two fail under a rule-based stagnation policy (Section~\ref{sec:ablation}). The adaptive gate addresses this by posterior-sampling over Beta arms, so the \texttt{guide} arm is down-weighted after a few unsuccessful injections, recovering \texttt{skip}-like behavior on mismatched tasks while retaining injection benefits elsewhere. Although the adaptive gate is not uniformly better than hand-tuned rules on benign tasks, it is the only variant in our study that remains safe across all five tasks without requiring per-task threshold tuning.

To evaluate content-level cross-task transfer, we use a \emph{Leave-One-Out (LOO) protocol}: when evaluating on task $k$, the memory bank excludes all experience from task $k$ itself.

We evaluate \ours{} on 8 benchmarks spanning circle packing, signal processing, load balancing, SQL optimization, and more, on two LLM backbones (GPT-5 and Gemini-3-Pro). Our key findings:

\begin{enumerate}
    \item \textbf{Consistent convergence improvement.} \ours{} achieves higher AUCC on \emph{all} 8 tasks ($p=0.0078$, Wilcoxon signed-rank) with $+8.7\%$ mean relative improvement on GPT-5 and $+11.7\%$ on Gemini-3-Pro.
    \item \textbf{Strongest early-stage acceleration.} On GPT-5, AUC@20 improves by $+9.4\%$ on average, with circle packing $+26.0\%$ and transaction scheduling $+27.2\%$, showing that the largest gains occur during cold-start convergence.
    \item \textbf{TS prevents catastrophic negative transfer.} Full \ours{} improves over AdaEvolve on all 5 ablation tasks; both na\"ive always-inject and rule-based stagnation variants fail to produce any score-improving program on 2/5 tasks.
    \item \textbf{Negligible overhead.} Cross-task transfer adds $<\!1\%$ per-iteration cost ($\sim$0.77\,s vs.\ 80--140\,s baseline); aggregate wall-clock time across the 8 main tasks is $0.69\times$ that of \ada{}.
\end{enumerate}

\section{Related Work}

\paragraph{LLM-based program evolution.}
Building on genetic programming~\citep{koza1992gp}, recent work replaces hand-designed variation operators with LLMs. FunSearch~\citep{romera2024funsearch}, AlphaEvolve~\citep{novikov2025alphaevolve}, and OpenEvolve~\citep{sharma2025openevolve} establish LLM-based evolutionary search on math and systems benchmarks; ShinkaEvolve~\citep{lange2025shinkaevolve}, CodeEvolve~\citep{assumpcao2025codeevolve}, GEPA~\citep{agrawal2025gepa}, and AdaEvolve~\citep{cemri2026adaevolve} extend these via sample-efficient sampling, island models, and adaptive paradigm breakthroughs. Earlier work includes ELM~\citep{lehman2022elm}, Language Model Crossover~\citep{meyerson2023crossover}, EoH~\citep{liu2024eoh}, OPRO~\citep{yang2024opro}, EvoPrompting~\citep{chen2023evoprompting}, and EvoX~\citep{liu2026evox}. \textbf{All operate per-task without explicit cross-task memory.} \ours{} addresses this gap by making cross-task transfer first-class at both the representation level (tactic memories) and the control level (adaptive intervention).

\paragraph{Selection in LLM-based program evolution.}
Recent LLM-evolution frameworks select \emph{within-task} programs for mutation through a variety of mechanisms. FunSearch~\citep{romera2024funsearch} and OpenEvolve~\citep{sharma2025openevolve} use island models with score-weighted parent sampling and periodic island resets; AlphaEvolve~\citep{novikov2025alphaevolve} extends this with multi-model collaboration; ShinkaEvolve~\citep{lange2025shinkaevolve} adds rejection-sampled adaptive parent sampling for sample efficiency; AdaEvolve~\citep{cemri2026adaevolve} couples adaptive island models with paradigm breakthroughs that periodically inject conceptually different strategies into the next mutation prompt. \textbf{All operate within one task: selection uses programs discovered for the current problem, and memorisable artifacts reset between tasks.} \ours{} instead adds a cross-task layer: a contextual Thompson Sampling gate~\citep{slivkins2019bandits,chapelle2011ts,kaufmann2012ts,agrawal2013ts} decides whether retrieved tactics should be withheld, suggested, or imposed. The closest precedent is within-run adaptive operator selection~\citep{dacosta2008dmab,fialho2010aos}; \ours{} instead gates cross-task interventions.

Related foundations include meta-learning~\citep{hospedales2022metalearning,finn2017maml}, retrieval-augmented code generation~\citep{lewis2020rag,lu2022reacc,parvez2021redcoder,zhou2023docprompting}, agentic memory~\citep{shinn2023reflexion,wang2023voyager,packer2023memgpt,park2023generative}, and continual learning~\citep{kirkpatrick2017ewc,rusu2016progressive,delange2021clsurvey}.

\section{Method}
\label{sec:method}

\subsection{Problem Formulation}

We consider a sequential setting where an LLM-based evolution system solves tasks $\mathcal{T} = \{T_1, \ldots, T_N\}$. Each task $T_i$ has description $d_i$, evaluation function $f_i: \mathcal{P} \to \mathbb{R}$, and seed program $p_i^0$; the system runs $M$ iterations per task to maximize $f_i$. \ours{} augments the standard (per-task) setting with a persistent memory state $\mathcal{M} = (\mathcal{B}, \pi)$ comprising a Tactic Memory Bank $\mathcal{B}$ and an Injection Policy $\pi$. The policy is updated during each task, while the bank is appended once at successful task completion; both are carried forward to subsequent tasks.

\subsection{Architecture Overview}

\begin{figure*}[t]
    \centering
    \includegraphics[width=\linewidth]{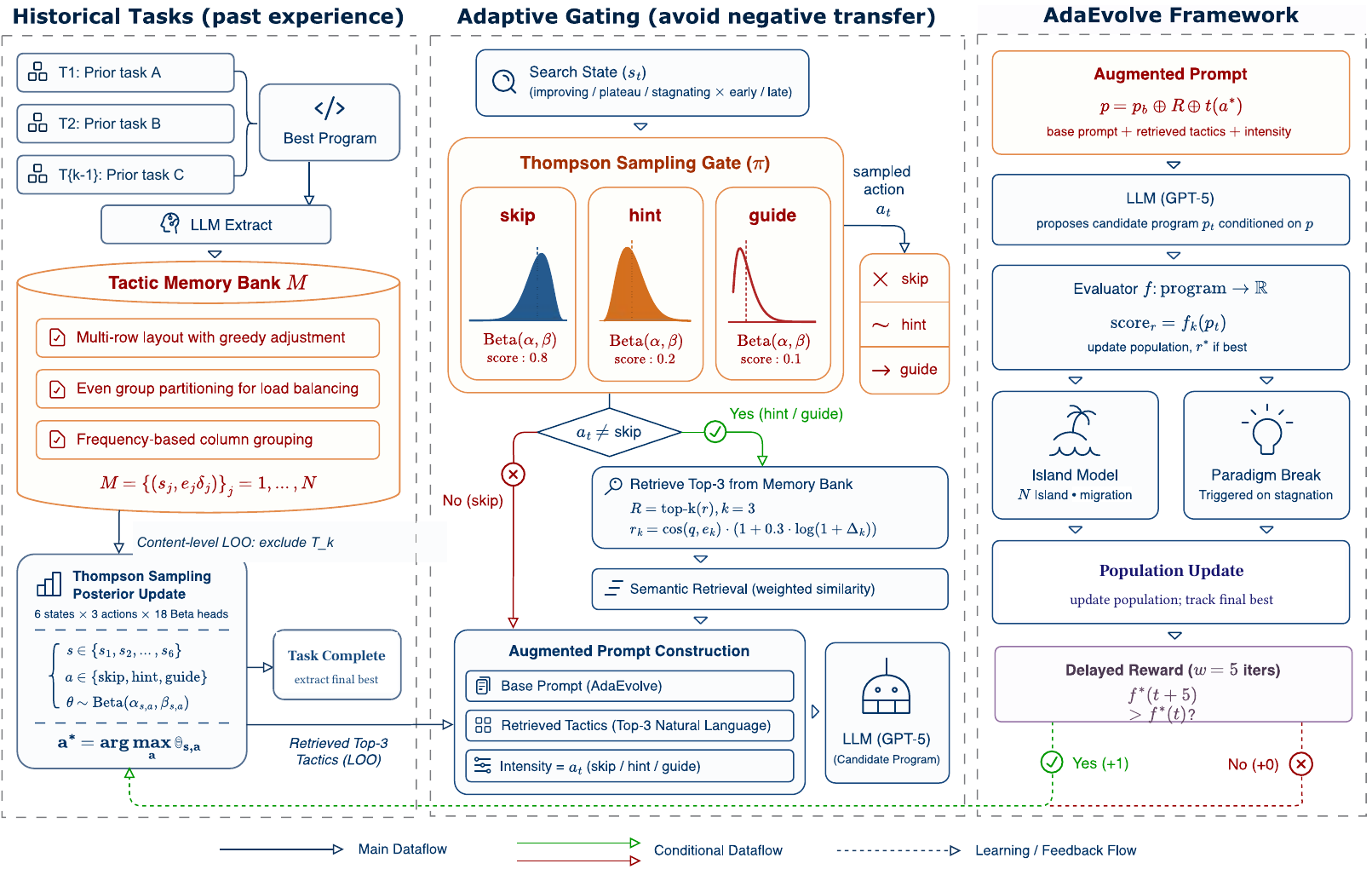}
    \caption{\textbf{\ours{} architecture.} Prior-task best programs are distilled into task-agnostic tactics under LOO. At each iteration, a state-conditioned Thompson Sampling gate selects \texttt{skip}, \texttt{hint}, or \texttt{guide}; retrieved tactics and the selected intensity augment the base prompt. AdaEvolve evaluates each candidate and returns a delayed reward that updates the gate. After a successful task run completes, its final best program contributes one tactic to memory for subsequent tasks.}
    \label{fig:architecture}
\end{figure*}

\ours{} wraps around the AdaEvolve~\citep{cemri2026adaevolve} base framework, adding two cross-task transfer modules (Figure~\ref{fig:architecture}): a \textbf{Tactic Memory Bank} $\mathcal{B}$ that stores strategy summaries from prior tasks (retrieved by embedding similarity), and an \textbf{Adaptive Injection Gate} $\pi$ that decides whether to inject memories and at what intensity. At iteration $t$ of task $T_i$, the controller (1) queries $\pi$ for an action $a_t \in \{\texttt{skip}, \texttt{hint}, \texttt{guide}\}$ given search state $s_t$; (2) if $a_t \neq \texttt{skip}$, retrieves the top-3 strategies from $\mathcal{B}_{-i}$; (3) constructs an augmented prompt at intensity $a_t$; (4) generates and evaluates a candidate; (5) resolves delayed rewards and updates $\pi$.

\subsection{Tactic Memory Bank}
\label{sec:memory_bank}

\paragraph{Strategy extraction.}
After a successful task run completes, we use GPT-4.1-mini to extract a task-agnostic strategy summary from its final best program:
\begin{equation}
    \text{tactic}_i = \text{LLM}_{\text{extract}}\!\left(d_i, \; p_i^{\text{best}}, \; f_i(p_i^{\text{best}})\right).
\end{equation}
The extraction prompt asks for general algorithmic descriptions (e.g., ``two-phase optimization with greedy initialization followed by gradient descent'') rather than task-specific code. Each completed successful task contributes one entry containing the tactic, approach-type label, task description, and embedding; the entry becomes available only to subsequent tasks.

\paragraph{Semantic retrieval.}
Given a new task with description $d_j$, we embed $\mathbf{e}_j = \text{Embed}(d_j)$ via text-embedding-3-small (1536-dim) and rank memory entries by a score-weighted similarity:
\begin{equation}
    r_k = \cos(\mathbf{e}_j, \mathbf{e}_k) \cdot \left(1 + 0.3 \cdot \log(1 + \max(\Delta_k, 0.01))\right),
\end{equation}
where $\Delta_k$ is the score improvement of entry $k$'s strategy over its baseline; we return the top-3.

\subsection{Adaptive Injection Gate}
\label{sec:policy}

The injection gate is implemented as a contextual bandit with Beta-distributed arms updated by Thompson sampling. We design it primarily as a \emph{gate against negative transfer} rather than a reward-maximizing controller: on tasks where retrieved strategies are mismatched, the posterior for \texttt{guide} collapses within a handful of unsuccessful injections and the policy reverts to predominantly \texttt{skip} actions.

\paragraph{State space.}
Six discrete states from two features: \textbf{search phase} $\in \{\texttt{improving}, \texttt{plateau}, \texttt{stagnating}\}$ from global improvement rate $\rho = \frac{\text{improvements}}{\text{evaluations}}$ ($\rho > 0.1 \Rightarrow \texttt{improving}$; $0.02 < \rho \le 0.1 \Rightarrow \texttt{plateau}$; $\rho \le 0.02 \Rightarrow \texttt{stagnating}$); and \textbf{task stage} $\in \{\texttt{early}, \texttt{late}\}$, where $t/M < 0.4$ is early and the remaining iterations are late.

\paragraph{Action space.}
Three injection intensities: \texttt{skip} (no injection, base prompt), \texttt{hint} (retrieved strategies as optional references), and \texttt{guide} (top strategy as recommended approach plus injection into paradigm breakthrough generation).

\paragraph{Informative priors.}
Each $(s,a)$ pair is $\text{Beta}(\alpha_{s,a}, \beta_{s,a})$. We set $\text{Prior}(\texttt{improving}, \texttt{skip}) = \text{Beta}(3,1)$ and $\text{Prior}(\texttt{stagnating}, \texttt{guide}) = \text{Beta}(3,1)$, with $\text{Beta}(1,1)$ for other pairs.

\paragraph{Decision and reward.}
At each iteration we sample $\theta_{s,a} \sim \text{Beta}(\alpha_{s,a}, \beta_{s,a})$ and pick $a^* = \arg\max_a \theta_{s,a}$. Before applying the action, we record the current best score $f^*_{\mathrm{pre}}$. The reward is delayed: after window $w=5$ iterations, $R_t = \mathbf{1}[f^*_{t+w} > f^*_{\mathrm{pre}} + \epsilon]$ with $\epsilon = 10^{-8}$; on success, $\alpha_{s_t,a_t} \mathrel{+}= 1$, otherwise $\beta_{s_t,a_t} \mathrel{+}= 1$. Concurrent pending decisions are resolved via a FIFO queue $\mathcal{Q}$. At the task boundary, unresolved records are evaluated against the final best score before the queue is cleared.

\subsection{Leave-One-Out Evaluation Protocol}
\label{sec:loo}

When evaluating \ours{} on task $T_k$: (1)~the memory bank $\mathcal{B}_{-k}$ excludes any entries from $T_k$; (2)~the policy $\pi$ retains its full posterior, since it learns task-agnostic state-action mappings rather than task-specific content or task identity; (3)~all methods share the same seed program and evaluator. Thus, the protocol leaves out target-task memory content while retaining the globally warm-started policy.

\begin{algorithm}[t]
\caption{\ours{}: One Iteration of Cross-Task Augmented Evolution}
\label{alg:memevo}
\begin{algorithmic}[1]
\REQUIRE Memory bank $\mathcal{B}_{-k}$, policy $\pi$ with Beta counts $\{(\alpha_{s,a}, \beta_{s,a})\}$ and pending queue $\mathcal{Q}$, iteration $t$, task description $d_k$, current best score $f^*_{t-1}$, reward window $w$, total iterations $M$
\STATE $s_t \gets \textsc{GetState}(\rho_t, t/M)$ \COMMENT{6-state discretisation}
\STATE Sample $\theta_{s_t,a} \sim \text{Beta}(\alpha_{s_t,a}, \beta_{s_t,a})$ for $a \in \{\texttt{skip}, \texttt{hint}, \texttt{guide}\}$
\STATE $a_t \gets \arg\max_{a} \theta_{s_t, a}$ \COMMENT{posterior sampling}
\STATE $f^*_{\mathrm{pre}} \gets f^*_{t-1}$;\quad $\text{tactics} \gets \emptyset$
\IF{$a_t \neq \texttt{skip}$}
    \STATE $\text{tactics} \gets \mathcal{B}_{-k}.\textsc{Retrieve}(d_k, \text{top\_k}{=}3)$
\ENDIF
\STATE $\text{prompt} \gets \textsc{BuildPrompt}(\text{parent}, \text{context}, \text{tactics}, a_t)$
\STATE $p_{\text{new}} \gets \text{LLM}(\text{prompt})$;\quad $\text{score} \gets f_k(p_{\text{new}})$
\STATE $f^*_t \gets \textsc{UpdatePopulation}(p_{\text{new}}, \text{score})$
\STATE $\mathcal{Q}.\textsc{Append}\bigl((s_t, a_t, t, f^*_{\mathrm{pre}})\bigr)$
\STATE $\textsc{ResolvePending}(\mathcal{Q}, t, w, f^*_t)$ \COMMENT{Beta updates for records with $t \ge \tau + w$}
\end{algorithmic}
\end{algorithm}

\section{Experiments}
\label{sec:experiments}

\subsection{Experimental Setup}

\paragraph{Benchmarks.} We evaluate 8 tasks across two categories: \emph{mathematical optimization} (4 from AlphaEvolve~\citep{novikov2025alphaevolve}: circle packing, signal processing, two inequality problems) and \emph{systems optimization} (4 from ADRS: EPLB, LLM-SQL, PRISM, transaction scheduling). All task objective scores are higher-is-better.

\paragraph{Methods.} We compare \textbf{\ours{}} (ours), which augments AdaEvolve with LOO memory and a TS gate; \textbf{\ada{}}, the base adaptive evolution framework; and \textbf{\topk{}}, a simpler top-$K$ mutation baseline.

\paragraph{Protocol.} GPT-5 via LiteLLM is the primary generator, with Gemini-3-Pro replication runs reported in the tables; text-embedding-3-small is used for retrieval. \textbf{Exp.~1} (memory accumulation): 8 tasks $\times$ 50 iterations $\times$ 2 methods, sequential. \textbf{Exp.~2} (main): 8 tasks $\times$ 80 iterations $\times$ 3 methods, using LOO memory from Exp.~1. \textbf{Exp.~3} (ablation): 5 tasks $\times$ 50 iterations $\times$ 5 variants. Each score is the mean over \textbf{5 paired seeds} $\{42, 0, 1, 2, 3\}$. Metrics are: \textbf{Final Score} at iteration 80; \textbf{AUCC} (normalized area under the convergence curve); \textbf{AUC@20} (cold-start efficiency); and \textbf{$T_{90\%}$} (iterations to 90\% of a method's own final score).

\subsection{Main Results}
\label{sec:main_results}

\begin{table*}[t]
\centering
\small
\begin{tabular}{@{}lrrrrrrrr@{}}
\toprule
& \multicolumn{4}{c}{\textbf{Backbone: GPT-5}} & \multicolumn{4}{c}{\textbf{Backbone: Gemini-3-Pro}} \\
\textbf{Task} & \ours{} & \ada{} & \topk{} & $\Delta$\% & \ours{} & \ada{} & \topk{} & $\Delta$\% \\
\midrule
circle\_packing & \textbf{0.9756} & 0.9524 & 0.9645 & +2.4\% & \textbf{0.9761} & 0.9534 & 0.9650 & +2.4\% \\
signal\_processing & \textbf{0.7227} & 0.7057 & 0.5508 & +2.4\% & \textbf{0.6685} & 0.5998 & 0.5095 & +11.5\% \\
first\_autocorr\_ineq & \textbf{0.9936} & 0.9913 & 0.9945 & +0.2\% & \textbf{0.9191} & 0.8426 & 0.9199 & +9.1\% \\
uncertainty\_ineq & \textbf{0.9024} & 0.8938 & 0.8975 & +1.0\% & \textbf{0.8347} & 0.7597 & 0.8302 & +9.9\% \\
eplb & \textbf{0.2171} & 0.1487 & 0.1336 & +46.0\% & \textbf{0.2260} & 0.1609 & 0.1391 & +40.5\% \\
llm\_sql & \textbf{0.7247} & 0.7047 & 0.7367 & +2.8\% & \textbf{0.7232} & 0.7019 & 0.7352 & +3.0\% \\
prism & 26.256 & \textbf{26.266} & 26.233 & $-$0.04\% & \textbf{26.230} & 26.213 & 26.207 & +0.06\% \\
txn\_scheduling & \textbf{4032.3} & 2777.8 & 2688.2 & +45.2\% & \textbf{3987.9} & 2716.0 & 2658.6 & +46.8\% \\
\bottomrule
\end{tabular}
\caption{\textbf{Final scores} (Exp.~2, 80 iter, mean over 5 seeds), on two LLM backbones. Bold: best of \ours{} / \ada{}.}
\label{tab:final_scores}
\end{table*}

On GPT-5, \ours{} achieves higher final scores than \ada{} on \textbf{7/8} tasks, with large gains on EPLB ($+46.0\%$) and txn\_scheduling ($+45.2\%$). The single loss on PRISM ($-0.04\%$) is negligible.

\subsection{Convergence Efficiency}

\begin{table*}[t]
\centering
\small
\begin{tabular}{@{}lrrr|rrr|rrr@{}}
\toprule
& \multicolumn{3}{c|}{\textbf{AUCC (full 80)}} & \multicolumn{3}{c|}{\textbf{AUC@20}} & \multicolumn{3}{c}{\textbf{$T_{90\%}$ (iter)}} \\
\textbf{Task} & \ours{} & \ada{} & $\Delta$ & \ours{} & \ada{} & $\Delta$ & \ours{} & \ada{} & Speedup \\
\midrule
\multicolumn{10}{@{}l}{\emph{Backbone: GPT-5}} \\
\midrule
circle\_pack & .946 & .885 & \textbf{+6.9\%} & .861 & .683 & \textbf{+26.0\%} & \textbf{3} & 9 & 3.0$\times$ \\
eplb & .159 & .135 & \textbf{+17.9\%} & .124 & .123 & +1.0\% & 53 & 50 & 0.9$\times$ \\
autocorr & .980 & .979 & +0.1\% & .942 & .942 & +0.0\% & 1 & 1 & 1.0$\times$ \\
llm\_sql & .708 & .696 & +1.8\% & .675 & .669 & +0.8\% & 1 & 1 & 1.0$\times$ \\
prism & 25.90 & 25.31 & +2.3\% & 24.84 & 22.45 & \textbf{+10.6\%} & \textbf{2} & 17 & 8.5$\times$ \\
signal & .668 & .651 & +2.5\% & .573 & .525 & \textbf{+9.2\%} & \textbf{15} & 23 & 1.5$\times$ \\
txn\_sched & 3766 & 2743 & \textbf{+37.3\%} & 3358 & 2639 & \textbf{+27.2\%} & 15 & 1$^\ddagger$ & --- \\
uncert\_ineq & .890 & .882 & +0.9\% & .855 & .848 & +0.8\% & 1 & 1 & 1.0$\times$ \\
\cmidrule(lr){2-10}
\textbf{Win/Total} & \multicolumn{3}{c|}{\textbf{8/8}} & \multicolumn{3}{c|}{\textbf{7/8}} & \multicolumn{3}{c}{\textbf{3/8}$^{\ddagger\ddagger}$} \\
\textbf{Mean $\Delta$} & \multicolumn{3}{c|}{\textbf{+8.7\%}} & \multicolumn{3}{c|}{\textbf{+9.4\%}} & & & \\
\midrule
\multicolumn{10}{@{}l}{\emph{Backbone: Gemini-3-Pro}} \\
\midrule
circle\_pack & .947 & .886 & \textbf{+6.9\%} & .861 & .684 & \textbf{+25.9\%} & \textbf{4} & 10 & 2.5$\times$ \\
eplb & .166 & .146 & \textbf{+13.3\%} & .129 & .133 & $-$3.0\% & 56 & 52 & 0.9$\times$ \\
autocorr & .907 & .832 & \textbf{+8.9\%} & .871 & .801 & \textbf{+8.8\%} & 2 & 2 & 1.0$\times$ \\
llm\_sql & .707 & .693 & +1.9\% & .674 & .666 & +1.1\% & 1 & 2 & 2.0$\times$ \\
prism & 25.87 & 25.26 & +2.4\% & 24.82 & 22.41 & \textbf{+10.8\%} & \textbf{3} & 16 & 5.3$\times$ \\
signal & .618 & .553 & \textbf{+11.7\%} & .530 & .446 & \textbf{+18.8\%} & \textbf{14} & 22 & 1.6$\times$ \\
txn\_sched & 3725 & 2683 & \textbf{+38.8\%} & 3321 & 2581 & \textbf{+28.7\%} & 16 & 1$^\ddagger$ & --- \\
uncert\_ineq & .823 & .750 & \textbf{+9.8\%} & .791 & .721 & \textbf{+9.7\%} & 2 & 2 & 1.0$\times$ \\
\cmidrule(lr){2-10}
\textbf{Win/Total} & \multicolumn{3}{c|}{\textbf{8/8}} & \multicolumn{3}{c|}{\textbf{7/8}} & \multicolumn{3}{c}{\textbf{4/8}$^{\ddagger\ddagger}$} \\
\textbf{Mean $\Delta$} & \multicolumn{3}{c|}{\textbf{+11.7\%}} & \multicolumn{3}{c|}{\textbf{+12.6\%}} & & & \\
\bottomrule
\end{tabular}

\begin{minipage}{\textwidth}
\footnotesize\raggedright
$^\ddagger$\ada{}'s $T_{90\%}{=}1$ when its final score is materially lower than \ours{}'s---reaching 90\% of a lower target is trivially faster.\\
$^{\ddagger\ddagger}$Excluding ties.
\end{minipage}
\caption{\textbf{Convergence efficiency} (\ours{} vs.\ \ada{} on 8 tasks, two backbones). $\Delta$\% = relative improvement.}
\label{tab:convergence}
\end{table*}

\paragraph{AUCC.} \ours{} wins on \textbf{all 8 tasks} (mean $+8.7\%$ on GPT-5, $+11.7\%$ on Gemini-3-Pro). Largest gains are on txn\_scheduling ($+37.3\%$/$+38.8\%$) and EPLB ($+17.9\%$/$+13.3\%$).

\paragraph{AUC@20.} Cold-start improves on \textbf{7/8 tasks} (mean $+9.4\%$/$+12.6\%$), exceeding the full-horizon AUCC gain and indicating that the largest improvements occur in the cold-start phase. The largest GPT-5 gains are circle\_packing ($+26.0\%$), txn\_scheduling ($+27.2\%$), and PRISM ($+10.6\%$).

\paragraph{$T_{90\%}$.} On GPT-5, \ours{} reaches 90\% faster on circle\_packing (3 vs.\ 9, $3.0\times$), PRISM (2 vs.\ 17, $8.5\times$), and signal\_processing (15 vs.\ 23, $1.5\times$).

\subsection{Statistical Significance}

On AUCC, the paired Wilcoxon signed-rank gives $p = 0.0078$ (the minimum two-sided $p$-value at $n{=}8$ when all task-level differences share the same sign); the bootstrap 95\% BCa CI on mean relative improvement is $[+2.9\%, +21.4\%]$; Cohen's $d_z = 0.675$ (medium effect).

\subsection{Ablation Study}
\label{sec:ablation}

We compare five variants on 5 tasks (50 iter): \textbf{\ours{} (full)} (LOO bank + TS), \textbf{MemEvo-always} (always guide), \textbf{MemEvo-stagnation} (rule-based: guide if stagnating, else skip), \textbf{\ada{}} (no memory), \textbf{\topk{}} (weak baseline).

\begin{table*}[t]
\centering
\small
\begin{tabular}{@{}lccccc@{}}
\toprule
\textbf{Task} & \textbf{\ours{}} & \textbf{always} & \textbf{stagnation} & \textbf{\ada{}} & \textbf{\topk{}} \\
\midrule
circle\_pack & 0.928 (+9.9\%) & 0.942 (+11.5\%) & \textbf{0.957} (+13.4\%) & 0.845 & 0.781 \\
eplb & \textbf{0.129} (+1.3\%) & \colorbox{red!15}{0.000} & \colorbox{red!15}{0.000} & 0.127 & 0.122 \\
signal & 0.639 (+2.9\%) & 0.599 ($-$3.5\%) & \textbf{0.646} (+4.0\%) & 0.621 & 0.421 \\
txn\_sched & 3647 (+34.0\%) & \textbf{4227} (+55.3\%) & 3548 (+30.3\%) & 2722 & --- \\
llm\_sql & \textbf{0.700} (+1.3\%) & \colorbox{red!15}{0.000} & \colorbox{red!15}{0.000} & 0.691 & 0.652 \\
\midrule
\textbf{Avg.\ $\Delta$\%} & \textbf{+9.9\%} & $-$27.4\% & $-$30.4\% & 0\% & $-$12.3\% \\
\textbf{Safe tasks} & \textbf{5/5} & 3/5 & 3/5 & 5/5 & 4/4 \\
\bottomrule
\end{tabular}
\caption{\textbf{Ablation}: AUCC@50 and relative improvement over \ada{}. \colorbox{red!15}{Red} = catastrophic failure (AUCC$=0$). \textbf{Avg.\ $\Delta$\%} averages per-task relative improvements over \ada{}; AUCC$=0$ failures are scored as $-100\%$.}
\label{tab:ablation}
\end{table*}

\paragraph{Both naive always-inject and rule-based stagnation fail catastrophically on mismatched tasks.}
MemEvo-always and MemEvo-stagnation both achieve AUCC${}=0$ on EPLB and LLM-SQL---\emph{no iteration in 50 produces a score-improving program} on these tasks, while \ours{} (full) and \ada{} both produce valid programs on the same tasks. MemEvo-always injects every iteration; MemEvo-stagnation injects whenever stagnation is detected. These failures are consistent with static policies repeatedly injecting inapplicable code patterns. TS limits this behavior because the \texttt{guide} posterior drops after unsuccessful injections, and the gate reverts to predominantly \texttt{skip}.

\paragraph{Memory injection is highly valuable when appropriate.}
On circle\_packing, signal, and txn\_scheduling, all three memory variants outperform \ada{} by $+2.9\%$ to $+55.3\%$, demonstrating genuine value when semantically aligned memories are available.

\paragraph{The adaptive gate trades peak upside for robustness.}
On circle\_packing ($+13.4\%$ vs.\ $+9.9\%$) and signal ($+4.0\%$ vs.\ $+2.9\%$), MemEvo-stagnation edges out \ours{} by small margins; on txn\_scheduling, MemEvo-always is best ($+55.3\%$). These results show that memory content is valuable when aligned, while the adaptive gate keeps that upside without exposing the system to static-injection brittleness. \ours{} is the only memory-augmented variant that improves over \ada{} on all 5 tasks without per-task threshold tuning.

\subsection{Analysis}
\label{sec:analysis}

\begin{figure}[t]
    \centering
    \includegraphics[width=0.98\columnwidth]{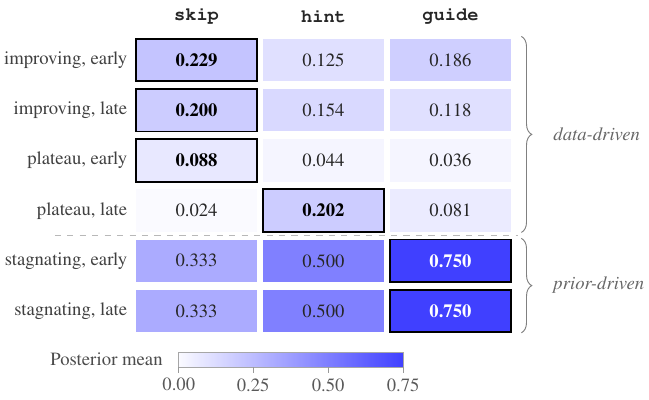}
    \caption{\textbf{Policy posterior} as a posterior-mean heatmap over 6 states $\times$ 3 actions after Exp.~1. Bold-bordered cells mark the greedy arm per state. Observed states yield the data-driven hierarchy improving $\to$ \texttt{skip}, plateau\_early $\to$ \texttt{skip}, and plateau\_late $\to$ \texttt{hint}; the unobserved \texttt{stagnating} states retain the prior-initialized \texttt{guide} action.}
    \label{fig:policy_heatmap}
\end{figure}

\paragraph{Policy posterior and prior sensitivity.}
Figure~\ref{fig:policy_heatmap} shows the warm-start policy posterior obtained from Exp.~1. The observed states recover an interpretable hierarchy: \textbf{improving $\to$ skip}, \textbf{plateau\_early $\to$ skip}, and \textbf{plateau\_late $\to$ hint}. These greedy actions are unchanged when the informative prior is replaced by uniform Beta(1,1), confirming that the observed-state behavior is learned rather than inherited from initialization. The unobserved \texttt{stagnating} states retain the informative prior that favors stronger memory use when progress stalls.

\paragraph{Reward rate validation.}
Within the \texttt{improving\_early} decisions in Exp.~2, \texttt{skip} attains a 39.6\% positive reward rate (99/250), compared with 34.0\% for \texttt{hint} (17/50) and 19.7\% for \texttt{guide} (15/76). \textbf{Over-injection during productive search actively harms performance}: \texttt{guide}'s reward rate is half that of \texttt{skip}.

\paragraph{Negative transfer avoided in practice.}
On EPLB, the closest retrievable memory (from LLM-SQL, cosine $0.481$) prescribes SQL-style grouping; under \texttt{MemEvo-always} this produces code that scores $0$ every iteration, whereas the adaptive gate's \texttt{guide}-arm posterior collapses after a handful of failures and the policy reverts to \texttt{skip}, allowing normal \ada{} search to proceed.

\paragraph{Retrieval forms semantic clusters.}
\label{sec:retrieval_analysis}
Retrieval similarity reveals two natural clusters: \emph{geometric placement} (similarities $0.420$--$0.646$) and \emph{numerical/systems} (similarities $0.375$--$0.555$). The catastrophic-failure pair (EPLB $\leftrightarrow$ LLM-SQL at $0.481$) is a \emph{post-hoc diagnostic}, not an inference-time cutoff: \ours{} always retrieves top-3 by weighted similarity and lets TS gate the decision.

\paragraph{Computational overhead.}
\ours{} adds $\sim$0.77\,s per iteration ($<\!1\%$ of the 80--140\,s baseline). Wall-clock time is faster on 5/8 main tasks, with an aggregate ratio of $0.69\times$ relative to \ada{}.

\begin{figure*}[t]
    \centering
    \includegraphics[width=\textwidth]{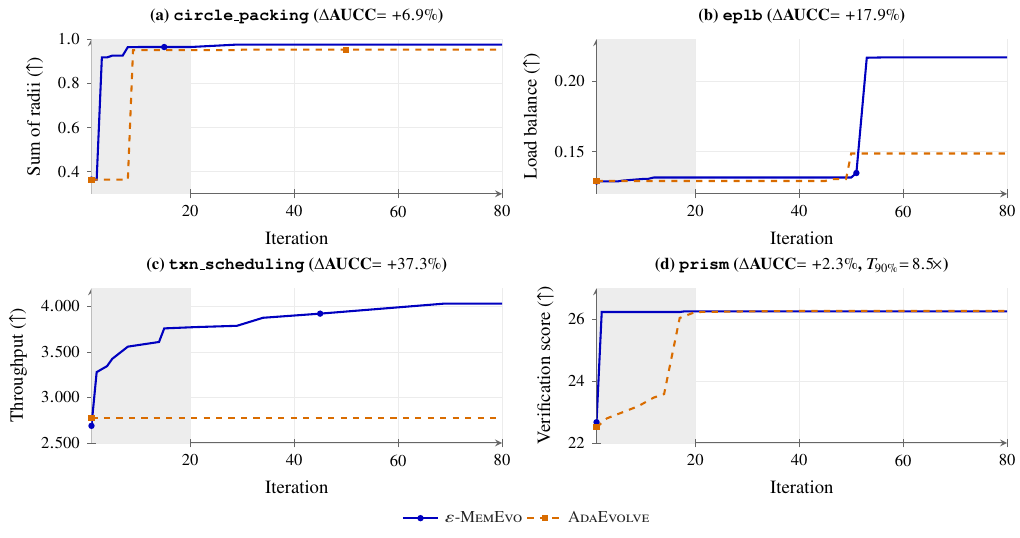}
    \caption{\textbf{Convergence curves} on four representative tasks. \ours{} (blue) converges faster than \ada{} (orange) in early iterations; the gap is most pronounced on circle\_packing and txn\_scheduling. Shaded region: AUC@20 zone.}
    \label{fig:convergence}
\end{figure*}

\section{Discussion}
\label{sec:discussion}

\paragraph{When does transfer help most?}
Our results suggest cross-task transfer is most beneficial when: (1)~the memory bank contains strategies from semantically related tasks (e.g., \texttt{circle\_packing} benefits from other geometric layout strategies at similarity $\ge 0.5$); (2)~the target task has a smooth fitness landscape where the base framework can make steady progress; (3)~the search is in its early stages (AUC@20 gains exceed full-horizon AUCC gains on both backbones). Transfer is less direct when available memories are from distant task domains, which is where the adaptive gate is most important.

\paragraph{Theoretical safety guarantee.}
Specialising the standard regret bound for Bernoulli Thompson sampling \citep{russo2018thompson, agrawal2013ts} to our 3-arm gating problem shows that, when retrieved tactics are mismatched (so $p_{\texttt{skip}} > p_{\texttt{guide}}$ in a fixed search state), the expected number of \texttt{guide} pulls in the first $T$ rounds satisfies $\mathbb{E}[T_{\texttt{guide}}(T)] \le C\log T / \Delta_{\texttt{guide}}^2 + o(\log T)$, so the per-round probability of injecting a harmful tactic decays as $O(1/t)$. In a synthetic validation over $T{=}500$ rounds, the rolling selection frequency of \texttt{guide} falls below 5\% by $t \approx 100$ in the mismatched setting.

\paragraph{Why Thompson sampling?}
Our gate needs to support delayed binary rewards, informative Beta priors, and rapid suppression of harmful actions. Fixed $\varepsilon$-greedy continues to select every arm with nonzero probability even after repeated failures, while UCB1 has no natural mechanism for incorporating the state-specific priors used by \ours{}. In matched and mismatched synthetic bandits, Thompson sampling is the only tested controller that combines near-lowest regret in both regimes with a vanishing harmful-arm selection rate. This choice does not imply that Thompson sampling is universally optimal: our LLM-in-the-loop experiments isolate adaptive gating against always-inject and stagnation-triggered controls, rather than replacing the gate with every bandit alternative.

\paragraph{Limitations.}
We evaluate two LLM backbones (GPT-5, Gemini-3-Pro). The memory bank is built at the task level: each entry distills a complete optimization run, and LOO evaluation tests transfer from other tasks. We do not yet know how retrieval precision, gate calibration, or negative-transfer frequency change as the bank grows to hundreds of entries. Moreover, \ours{} augments rather than replaces its base evolution framework, so memory transfer cannot compensate for fundamental exploration failures in the underlying search. A stronger rule-based control not evaluated here would combine stagnation-triggered injection with a $k$-strike auto-disable rule. Such a controller could recover much of the observed safety benefit without posterior sampling, but would introduce a task-dependent strike threshold; comparing it with \ours{} is an important missing ablation. The within-paper \ada{} vs.\ \ours{} comparison (Tables~\ref{tab:final_scores}--\ref{tab:convergence}) remains the primary controlled test.

\section{Conclusion}
\label{sec:conclusion}

We presented \ours{}, a framework that formalizes cross-task knowledge transfer as an \emph{adaptive intervention problem} in LLM-based program evolution. Through a Tactic Memory Bank for content-level transfer and an Adaptive Injection Gate for strategy-level transfer, \ours{} achieves statistically significant convergence improvements across 8 diverse optimization benchmarks on GPT-5 ($p=0.0078$, 8/8 AUCC wins, $+8.7\%$ average improvement). The key insight is that \emph{adaptive} injection control is essential: while cross-task memories can provide substantial value (up to $+55\%$), uncontrolled injection can actively harm the search---in our ablation, two of five tasks fail to improve under naive always-inject. The adaptive gate resolves this tension with an interpretable posterior: data favor \emph{skip} in improving and early-plateau states and \emph{hint} in late-plateau states, while the unobserved stagnating states retain the prior-initialized \emph{guide} action. This control incurs negligible overhead ($<\!1\%$). The content-level LOO protocol excludes target-task entries from the memory bank, so the transferred tactics come from other tasks rather than target solutions.

These results establish adaptive cross-task memory as a complementary layer for evolutionary coding agents. Next steps include larger, more diverse banks, hierarchical memory, and transfer across base search frameworks and weaker LLM backbones.

\subsection*{Broader Impact Statement}

\ours{} uses $0.69\times$ the aggregate wall-clock time of \ada{} across the 8 main tasks but costs 2--34 GPU-hours per task; scaling may raise energy use. Cross-task reuse lowers cost; gating reveals memory influence.

\bibliography{references}

@article{romera2024funsearch,
  title={Mathematical discoveries from program search with large language models},
  author={Romera-Paredes, Bernardino and Barekatain, Mohammadamin and Novikov, Alexander and Balog, Matej and Kumar, M. Pawan and Dupont, Emilien and Ruiz, Francisco J. R. and Ellenberg, Jordan and Wang, Pengming and Fawzi, Omar and Kohli, Pushmeet and Fawzi, Alhussein},
  journal={Nature},
  volume={625},
  pages={468--475},
  year={2024},
  doi={10.1038/s41586-023-06924-6}
}

@article{novikov2025alphaevolve,
  title={{AlphaEvolve}: A coding agent for scientific and algorithmic discovery},
  author={Novikov, Alexander and Vu, Ngan and Eisenberger, Marvin and Dupont, Emilien and Huang, Po-Sen and Wagner, Adam Zsolt and others},
  journal={arXiv preprint arXiv:2506.13131},
  year={2025}
}

@article{cemri2026adaevolve,
  title={{AdaEvolve}: Adaptive {LLM} Driven Zeroth-Order Optimization},
  author={Cemri, Mert and Agrawal, Shubham and Gupta, Akshat and Liu, Shu and Cheng, Audrey and Mang, Qiuyang and Naren, Ashwin and Erdogan, Lutfi Eren and Sen, Koushik and Zaharia, Matei and Dimakis, Alex and Stoica, Ion},
  journal={arXiv preprint arXiv:2602.20133},
  year={2026}
}

@inproceedings{chen2023evoprompting,
  title={{EvoPrompting}: Language Models for Code-Level Neural Architecture Search},
  author={Chen, Angelica and Dohan, David M. and So, David R.},
  booktitle={Advances in Neural Information Processing Systems},
  year={2023}
}

@article{lehman2022elm,
  title={Evolution through Large Models},
  author={Lehman, Joel and Gordon, Jonathan and Jain, Shawn and Ndousse, Kamal and Yeh, Cathy and Stanley, Kenneth O.},
  journal={arXiv preprint arXiv:2206.08896},
  year={2022}
}

@inproceedings{liu2024eoh,
  title={Evolution of Heuristics: Towards Efficient Automatic Algorithm Design Using Large Language Model},
  author={Liu, Fei and Tong, Xialiang and Yuan, Mingxuan and Lin, Xi and Luo, Fu and Wang, Zhenkun and Lu, Zhichao and Zhang, Qingfu},
  booktitle={International Conference on Machine Learning},
  year={2024}
}

@inproceedings{yang2024opro,
  title={Large Language Models as Optimizers},
  author={Yang, Chengrun and Wang, Xuezhi and Lu, Yifeng and Liu, Hanxiao and Le, Quoc V. and Zhou, Denny and Chen, Xinyun},
  booktitle={International Conference on Learning Representations},
  year={2024}
}

@article{liu2026evox,
  title={{EvoX}: Meta-Evolution for Automated Discovery},
  author={Liu, Shu and Agarwal, Shubham and Maheswaran, Monishwaran and Cemri, Mert and others},
  journal={arXiv preprint arXiv:2602.23413},
  year={2026}
}

@article{meyerson2023crossover,
  title={Language Model Crossover: Variation through Few-Shot Prompting},
  author={Meyerson, Elliot and Nelson, Mark J. and Bradley, Herbie and Gaier, Adam and Moradi, Arash and Hoover, Amy K. and Lehman, Joel},
  journal={arXiv preprint arXiv:2302.12170},
  year={2023}
}

@article{russo2018thompson,
  title={A Tutorial on {Thompson} Sampling},
  author={Russo, Daniel and Van Roy, Benjamin and Kazerouni, Abbas and Osband, Ian and Wen, Zheng},
  journal={Foundations and Trends in Machine Learning},
  volume={11},
  number={1},
  pages={1--96},
  year={2018}
}

@article{slivkins2019bandits,
  title={Introduction to Multi-Armed Bandits},
  author={Slivkins, Aleksandrs},
  journal={Foundations and Trends in Machine Learning},
  volume={12},
  number={1--2},
  pages={1--286},
  year={2019}
}

@inproceedings{lewis2020rag,
  title={Retrieval-Augmented Generation for Knowledge-Intensive {NLP} Tasks},
  author={Lewis, Patrick and Perez, Ethan and Piktus, Aleksandra and Petroni, Fabio and Karpukhin, Vladimir and Goyal, Naman and K{\"u}ttler, Heinrich and Lewis, Mike and Yih, Wen-tau and Rockt{\"a}schel, Tim and Riedel, Sebastian and Kiela, Douwe},
  booktitle={Advances in Neural Information Processing Systems},
  year={2020}
}

@inproceedings{shinn2023reflexion,
  title={Reflexion: Language Agents with Verbal Reinforcement Learning},
  author={Shinn, Noah and Cassano, Federico and Gopinath, Ashwin and Narasimhan, Karthik and Yao, Shunyu},
  booktitle={Advances in Neural Information Processing Systems},
  year={2023}
}

@inproceedings{wang2023voyager,
  title={Voyager: An Open-Ended Embodied Agent with Large Language Models},
  author={Wang, Guanzhi and Xie, Yuqi and Jiang, Yunfan and Mandlekar, Ajay and Xiao, Chaowei and Zhu, Yuke and Fan, Linxi and Anandkumar, Anima},
  booktitle={Advances in Neural Information Processing Systems},
  year={2023}
}

@article{hospedales2022metalearning,
  title={Meta-Learning in Neural Networks: A Survey},
  author={Hospedales, Timothy and Antoniou, Antreas and Micaelli, Paul and Storkey, Amos},
  journal={IEEE Transactions on Pattern Analysis and Machine Intelligence},
  volume={44},
  number={9},
  pages={5149--5169},
  year={2022}
}

@book{koza1992gp,
  title={Genetic Programming: On the Programming of Computers by Means of Natural Selection},
  author={Koza, John R.},
  publisher={MIT Press},
  year={1992}
}

@misc{sharma2025openevolve,
  title={{OpenEvolve}: An Open-Source Evolutionary Coding Agent},
  author={Sharma, Asankhaya},
  year={2025},
  howpublished={\url{https://github.com/algorithmicsuperintelligence/openevolve}}
}

@article{agrawal2025gepa,
  title={{GEPA}: Reflective Prompt Evolution Can Outperform Reinforcement Learning},
  author={Agrawal, Lakshya A. and Tan, Shangyin and Soylu, Dilara and Ziems, Noah and Khare, Rishi and Opsahl-Ong, Krista and Singhvi, Arnav and Shandilya, Herumb and Ryan, Michael J. and Jiang, Meng and others},
  journal={arXiv preprint arXiv:2507.19457},
  year={2025}
}

@article{lange2025shinkaevolve,
  title={{ShinkaEvolve}: Towards Open-Ended and Sample-Efficient Program Evolution},
  author={Lange, Robert Tjarko and Imajuku, Yuki and Cetin, Edoardo},
  journal={arXiv preprint arXiv:2509.19349},
  year={2025}
}

@article{assumpcao2025codeevolve,
  title={{CodeEvolve}: An Open Source Evolutionary Coding Agent for Algorithm Discovery and Optimization},
  author={Assump{\c{c}}{\~a}o, Henrique and Ferreira, Diego and Campos, Leonardo and Murai, Fabricio},
  journal={arXiv preprint arXiv:2510.14150},
  year={2025}
}

@inproceedings{lu2022reacc,
  title={{ReACC}: A Retrieval-Augmented Code Completion Framework},
  author={Lu, Shuai and Duan, Nan and Han, Hojae and Guo, Daya and Hwang, Seung-won and Svyatkovskiy, Alexey},
  booktitle={Proceedings of the 60th Annual Meeting of the Association for Computational Linguistics},
  year={2022}
}

@inproceedings{parvez2021redcoder,
  title={Retrieval Augmented Code Generation and Summarization},
  author={Parvez, Md Rizwan and Ahmad, Wasi Uddin and Chakraborty, Saikat and Ray, Baishakhi and Chang, Kai-Wei},
  booktitle={Findings of the Association for Computational Linguistics: EMNLP 2021},
  year={2021}
}

@inproceedings{zhou2023docprompting,
  title={{DocPrompting}: Generating Code by Retrieving the Docs},
  author={Zhou, Shuyan and Alon, Uri and Xu, Frank F. and Wang, Zhiruo and Jiang, Zhengbao and Neubig, Graham},
  booktitle={International Conference on Learning Representations},
  year={2023}
}

@article{packer2023memgpt,
  title={{MemGPT}: Towards {LLMs} as Operating Systems},
  author={Packer, Charles and Wooders, Sarah and Lin, Kevin and Fang, Vivian and Patil, Shishir G. and Stoica, Ion and Gonzalez, Joseph E.},
  journal={arXiv preprint arXiv:2310.08560},
  year={2023}
}

@inproceedings{park2023generative,
  title={Generative Agents: Interactive Simulacra of Human Behavior},
  author={Park, Joon Sung and O'Brien, Joseph C. and Cai, Carrie J. and Morris, Meredith Ringel and Liang, Percy and Bernstein, Michael S.},
  booktitle={Proceedings of the 36th Annual ACM Symposium on User Interface Software and Technology (UIST)},
  year={2023}
}

@inproceedings{finn2017maml,
  title={Model-Agnostic Meta-Learning for Fast Adaptation of Deep Networks},
  author={Finn, Chelsea and Abbeel, Pieter and Levine, Sergey},
  booktitle={International Conference on Machine Learning},
  year={2017}
}

@article{kirkpatrick2017ewc,
  title={Overcoming catastrophic forgetting in neural networks},
  author={Kirkpatrick, James and Pascanu, Razvan and Rabinowitz, Neil and Veness, Joel and Desjardins, Guillaume and Rusu, Andrei A. and Milan, Kieran and Quan, John and Ramalho, Tiago and Grabska-Barwinska, Agnieszka and Hassabis, Demis and Clopath, Claudia and Kumaran, Dharshan and Hadsell, Raia},
  journal={Proceedings of the National Academy of Sciences},
  volume={114},
  number={13},
  pages={3521--3526},
  year={2017}
}

@article{rusu2016progressive,
  title={Progressive Neural Networks},
  author={Rusu, Andrei A. and Rabinowitz, Neil C. and Desjardins, Guillaume and Soyer, Hubert and Kirkpatrick, James and Kavukcuoglu, Koray and Pascanu, Razvan and Hadsell, Raia},
  journal={arXiv preprint arXiv:1606.04671},
  year={2016}
}

@article{delange2021clsurvey,
  title={A Continual Learning Survey: Defying Forgetting in Classification Tasks},
  author={De Lange, Matthias and Aljundi, Rahaf and Masana, Marc and Parisot, Sarah and Jia, Xu and Leonardis, Ale{\v{s}} and Slabaugh, Gregory and Tuytelaars, Tinne},
  journal={IEEE Transactions on Pattern Analysis and Machine Intelligence},
  volume={44},
  number={7},
  pages={3366--3385},
  year={2021}
}

@inproceedings{chapelle2011ts,
  title={An Empirical Evaluation of {T}hompson Sampling},
  author={Chapelle, Olivier and Li, Lihong},
  booktitle={Advances in Neural Information Processing Systems (NeurIPS)},
  year={2011}
}

@inproceedings{kaufmann2012ts,
  title={{T}hompson Sampling: An Asymptotically Optimal Finite-Time Analysis},
  author={Kaufmann, Emilie and Korda, Nathaniel and Munos, R{\'e}mi},
  booktitle={Algorithmic Learning Theory (ALT)},
  year={2012}
}

@inproceedings{agrawal2013ts,
  title={Further Optimal Regret Bounds for {T}hompson Sampling},
  author={Agrawal, Shipra and Goyal, Navin},
  booktitle={International Conference on Artificial Intelligence and Statistics (AISTATS)},
  year={2013}
}

@inproceedings{dacosta2008dmab,
  title={Adaptive Operator Selection with Dynamic Multi-Armed Bandits},
  author={Da Costa, Lu{\'\i}s and Fialho, {\'A}lvaro and Schoenauer, Marc and Sebag, Mich{\`e}le},
  booktitle={Genetic and Evolutionary Computation Conference (GECCO)},
  year={2008}
}

@article{fialho2010aos,
  title={Analyzing Bandit-based Adaptive Operator Selection Mechanisms},
  author={Fialho, {\'A}lvaro and Da Costa, Lu{\'\i}s and Schoenauer, Marc and Sebag, Mich{\`e}le},
  journal={Annals of Mathematics and Artificial Intelligence},
  volume={60},
  number={1-2},
  pages={25--64},
  year={2010}
}

\end{document}